\documentclass{article}
\usepackage{spconf,amsmath,graphicx}

\usepackage{hyperref}
\usepackage{xcolor}

\usepackage{enumitem}
\setlist{nosep, leftmargin=14pt}

\usepackage{mwe} 
\usepackage{multirow}

\newif\ifshowedits
\showeditsfalse   

\newcommand{\uma}[1]{\ifshowedits\textcolor{blue}{#1}\else#1\fi}

\newcommand{\JJN}[1]{\ifshowedits\textcolor{cyan}{#1}\else#1\fi}

\definecolor{mypink}{RGB}{219,48,122}

\title{Uncertainty-Aware Image Classification In Biomedical Imaging Using Spectral-normalized Neural Gaussian Processes}
\name{Uma Meleti, Jeffrey J. Nirschl}
\address{Department of Pathology and Lab Medicine, University of Wisconsin-Madison}
\begin{document}
%
\maketitle
\begin{abstract}
Accurate histopathologic interpretation is key for clinical decision-making; however, current deep learning models for digital pathology are often overconfident and poorly calibrated in out-of-distribution (OOD) settings, which limit trust and clinical adoption. Safety-critical medical imaging workflows benefit from intrinsic uncertainty-aware properties that can accurately reject OOD input. We implement the Spectral-normalized Neural Gaussian Process (SNGP), a set of lightweight modifications that apply spectral normalization and replace the final dense layer with a Gaussian process layer to improve single-model uncertainty estimation and OOD detection. We evaluate SNGP vs. deterministic and Monte Carlo dropout on six datasets across three biomedical classification tasks: white blood cells, amyloid plaques, and colorectal histopathology. SNGP has comparable in-distribution performance while significantly improving uncertainty estimation and OOD detection. Thus, SNGP or related models offer a useful framework for uncertainty-aware classification in digital pathology, supporting safe deployment and building trust with pathologists.

\end{abstract}

\begin{keywords}
Digital pathology, uncertainty quantification, out-of-distribution detection, SNGP
\end{keywords}
\section{Introduction}
\label{sec:intro}
\begin{figure}[!t]
    \centering
    \includegraphics[width=0.8\columnwidth, height=0.28\textheight]{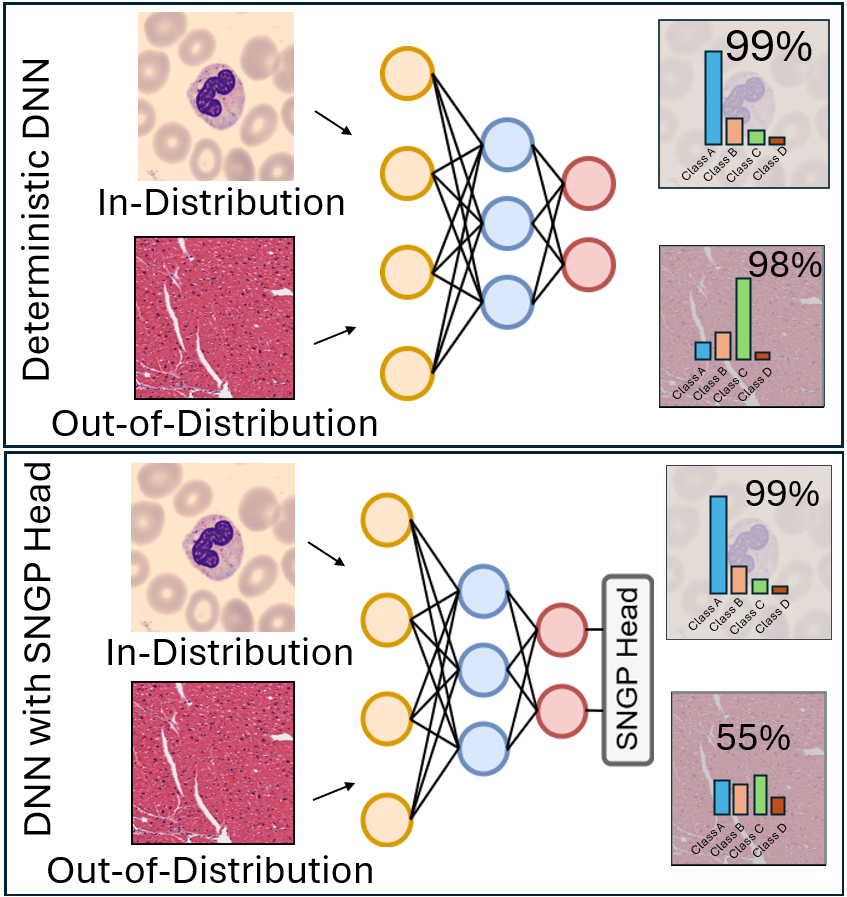}
    \vspace{-5pt}
    \caption{Overview of the predictions of deterministic DNN vs SNGP on In-Distribution (blood cell) vs OOD (cardiac) data.}
    \label{fig:overview}
    \vspace{-5pt}
\end{figure}

Digital pathology is transforming biomedical imaging by enabling quantitative analysis of tissue architecture at scale. In clinical practice, pathology interpretation by trained experts remains the gold standard for diagnosis in oncology and neurodegenerative diseases. Despite recent advances in deep learning, healthcare systems are cautious and slow to integrate artificial intelligence (AI) into diagnostic workflows due to logistical, regulatory, and safety reasons. One factor is that many deep neural networks (DNN) are often overconfident \cite{guo2017_dnn_calib} and lack intrinsic uncertainty-aware properties, assigning high softmax probabilities to OOD inputs. Clinicians can misinterpret the softmax probability as a reliable confidence metric, even though it is generally poorly calibrated, and experience disappointment with obviously incorrect high-probability predictions on new data. This erodes pathologists' trust and poses safety risks in the medical domain, where errors can result in patient harm.


Safe deployment of AI systems requires both accurate and uncertainty-aware predictions. Ideally, models with uncertainty-aware properties will have predictive probabilities that, in part, reflect the distance of the input to the training data manifold and propagate that uncertainty to the clinician. Existing approaches to uncertainty quantification, including Bayesian neural networks, Monte Carlo (MC) dropout, and deep ensembles, are potential approaches, but they are computationally expensive, requiring multiple stochastic forward passes. These methods are therefore difficult to scale to whole-slide images and impractical for real-time clinical use.

Recently, Spectral-normalized Neural Gaussian Processes (SNGP) have been proposed as an efficient approximation to Bayesian inference. 
SNGP models propagate predictive uncertainty with a single forward pass. This property makes SNGP particularly well-suited for computational pathology, where both efficiency and reliability are essential.

In this study, we implement and evaluate SNGP for uncertainty-aware image classification across multiple biomedical domains. Using six datasets across three tasks (white blood cell, amyloid plaque, and colorectal histopathology), we compare SNGP to deterministic and MC dropout baselines. SNGP maintains high in-distribution accuracy while improving \JJN{ OOD detection and uncertainty-aware behavior.}

\uma{Our contributions are threefold. First, we provide a systematic evaluation of SNGP across diverse pathology datasets. Second, we compare with deterministic and Monte Carlo methods for uncertainty quantification. Third, we release an open-source PyTorch SNGP framework to support reproducible uncertainty-aware deep learning in pathology. Together, these contributions provide a rigorous empirical assessment of SNGP in digital pathology and practical guidance for deploying trustworthy AI systems.}




\section{Related work}
\label{sec:rel work}

Deep learning models for digital pathology are typically deterministic, producing single predictions without calibrated confidence estimates. Bayesian neural networks provide a formal approach to uncertainty estimation but are computationally impractical for large architectures. Approximate methods such as Monte Carlo (MC) dropout \cite{gal2016_dropout_bayesian} and deep ensembles \cite{lakshminarayanan2017_deep_ensembles} improve reliability but require multiple stochastic forward passes, limiting scalability for high-resolution whole-slide analysis. Deterministic alternatives, including evidential and distributionally robust models, have been explored but often train unstably or scale poorly.

Model calibration and OOD detection are important for clinical reliability. Metrics such as the Expected Calibration Error (ECE) and Brier score assess the alignment between model confidence and accuracy, while OOD detection methods based on Mahalanobis distance, energy scoring, or temperature scaling can flag unfamiliar inputs. These techniques improve robustness but require more compute, whereas DNNs with built-in uncertainty-aware properties offer simplicity and interpretability in the clinical setting.

In biomedical imaging, deep learning has achieved strong results for white blood cell classification \cite{Acevedo2020-ja, jung2022wbc}, cardiac pathology \cite{Nirschl2018-pc}, amyloid plaque classification \cite{Tang2019-fy, Wong2022-nm}, and colorectal histopathology \cite{Kather2016-db, LozanoNirschl2024_microbench}. Yet, models often fail to generalize across institutions due to differences in staining, scanners, and sample preparation. Most provide uncalibrated confidence scores, limiting interpretability and clinical trust.


\section{Methods and Experiments}

\begin{figure}
    \centering
    \newcommand{\figwidth}{0.30\columnwidth}
    \newcommand{\figheight}{0.10\textheight}

    \setlength{\tabcolsep}{0pt} 
    \renewcommand{\arraystretch}{0.6}

    \begin{tabular}{ccc}
    \includegraphics[width=\figwidth, height=\figheight]{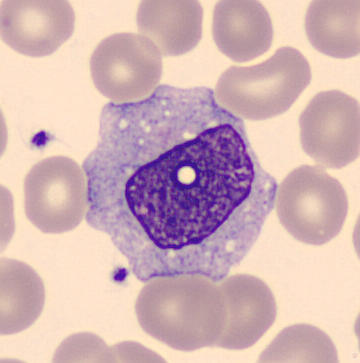} &
    \includegraphics[width=\figwidth, height=\figheight]{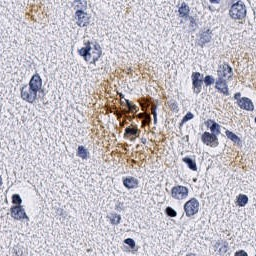} &
    \includegraphics[width=\figwidth, height=\figheight]{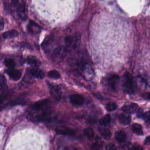} \\
    (a) Acevedo & (b) Tang & (c) Kather '16 \\[2pt]
    \includegraphics[width=\figwidth, height=\figheight]{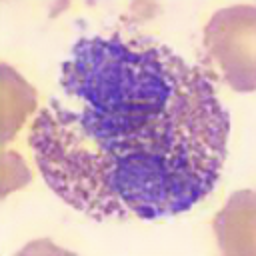} &
    \includegraphics[width=\figwidth, height=\figheight]{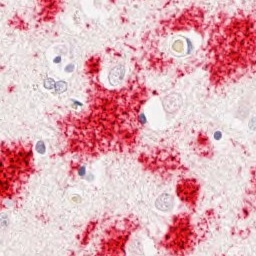} &

    \includegraphics[width=\figwidth, height=\figheight]{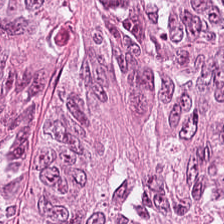} \\
    (d) Jung & (e) Wong & (f) Kather '18
    \end{tabular}
    
    \vspace{-5pt}
    \caption{Sample images from paired datasets: (a,d) white blood cells, (b,e) amyloid plaques, and (c,f) colorectal pathology.}
    \label{fig:dataset_examples}
    \vspace{-5pt}
\end{figure}

\subsection{Datasets}

All datasets are publicly available from the original authors or MicroBench \cite{LozanoNirschl2024_microbench}. Original train/val/test splits were used when available; otherwise, stratified class-based splits were created. The data cover three representative histopathology classification tasks spanning tissue, stain, pathology specialty, and institutional source, enabling ID and OOD evaluation.

White blood cell (WBC) datasets include Acevedo et al. (2020) and Jung et al. (2022), which contain peripheral blood smear images acquired from different laboratories and microscopes.
Amyloid plaque detection uses datasets from Tang et al. (2019) and Wong et al. (2022), consisting of immunostained human brain sections with amyloid pathology labeled by expert neuropathologists.
Colorectal histopathology tasks use the Kather et al. (2016, 2018) datasets of hematoxylin and eosin–stained tissue patches representing normal epithelium, stroma, and multiple carcinoma subtypes.

Each dataset \uma{followed} MicroBench preprocessing: RGB normalization to zero mean and unit variance, resizing to 224×224 pixels, and stratified train/test splits. No new data were generated.
Dataset pairs (Fig. \ref{fig:dataset_examples}) representing the same biological task but 2 distinct acquisition domains 
\uma{were used for OOD evaluation to assess robustness under domain shift.}

\begin{figure*}[!t]
  \centering
  \includegraphics[width=2.0\columnwidth, height=0.20\textheight]{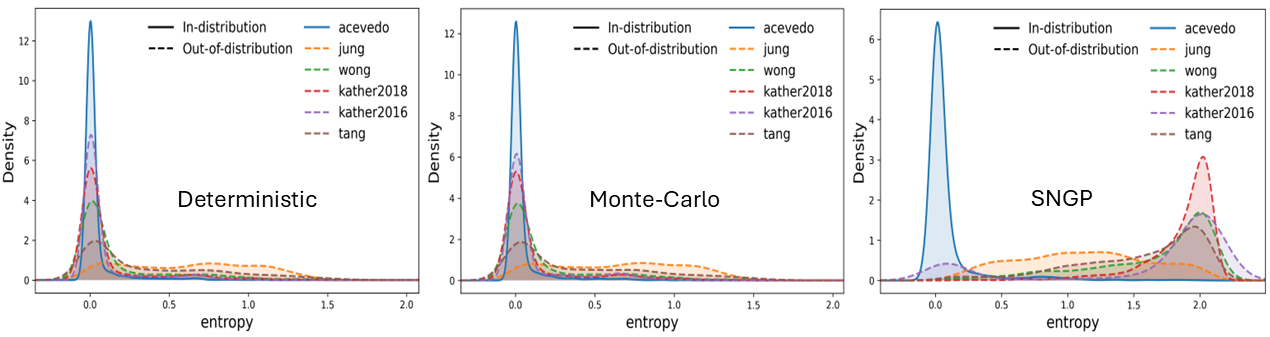}
  \vspace{-5pt}
  \JJN{\caption{\textbf{Class-logit entropy distributions for ID and OOD samples.} Deterministic and MC Dropout models show low entropy on OOD data, indicating overconfident predictions. In contrast, SNGP assigns higher entropy for OOD inputs, reflecting better uncertainty awareness and separation between ID and OOD inputs.}}
  \label{fig:entropy}
  \vspace{-5pt}
\end{figure*}

\subsection{Spectral Normalized Gaussian Process (SNGP)}
\label{sec: sngp}


SNGP integrates spectral normalization into the hidden layers to control the Lipschitz constant and stabilize feature representations. It replaces the conventional final layer with a Gaussian Process (GP) classifier approximated using Random Fourier Features (RFF), enabling scalable GP inference. The RFF layer maps the normalized features into an approximate kernel space, allowing closed-form computation of the posterior mean and variance, which naturally increase as inputs drift away from the training distribution. A detailed implementation of SNGP is described in detail in \cite{liu2020_sngp}. Fig. \ref{fig:overview} shows how the output probabilities are distributed between the ID and OOD inputs for deterministic and SNGP model.


\begin{table*}[t]
\centering
\caption{Comparison of baseline, MC Dropout, and SNGP trained on Acevedo dataset and tested across other datasets.}
\label{tab:results_acevedo}
\setlength{\tabcolsep}{5pt}
\renewcommand{\arraystretch}{1.0}
\begin{tabular}{lcccccc}
\hline
\textbf{Method} & \textbf{Jung} & \textbf{Wong} & \textbf{Kather2016} & \textbf{Kather2018} & \textbf{Nirschl} & \textbf{Tang} \\
\hline
Baseline & 0.957 $\pm$ 0.003 & 0.613 $\pm$ 0.019 & 0.488 $\pm$ 0.014 & 0.394 $\pm$ 0.012 & 0.386 $\pm$ 0.014 & 0.831 $\pm$ 0.006 \\
MC Dropout & 0.959 $\pm$ 0.002 & 0.622 $\pm$ 0.019 & 0.526 $\pm$ 0.013 & 0.405 $\pm$ 0.012 & 0.495 $\pm$ 0.013 & 0.837 $\pm$ 0.006 \\
SNGP & \textbf{0.981 $\pm$ 0.002} & \textbf{0.992 $\pm$ 0.001} & \textbf{0.971 $\pm$ 0.002} & \textbf{0.998 $\pm$ 0.001} & \textbf{1.000 $\pm$ 0.000} & \textbf{0.992 $\pm$ 0.001} \\
\hline
FID & 306.65 & 351.25 & 375.57 & 377.93 & 379.89 & 388.94 \\
\uma{Dataset Size} & 5000 & 70044 & 5000 & 100000 & 2299 & 80873 \\
\hline
\end{tabular}
\end{table*}

\begin{table*}[t]
\centering
\caption{Comparison of baseline, MC Dropout, and SNGP trained on Wong dataset and tested across other datasets.}
\label{tab:results_wong}
\setlength{\tabcolsep}{5pt}
\renewcommand{\arraystretch}{1.0}
\begin{tabular}{lcccccc}
\hline
\textbf{Method} & \textbf{Tang} & \textbf{Kather2016} & \textbf{Kather2018} & \textbf{Jung} & \textbf{Nirschl} & \textbf{Acevedo} \\
\hline
Baseline & 0.464 $\pm$ 0.014 & 0.703 $\pm$ 0.010 & 0.664 $\pm$ 0.016 & 0.497 $\pm$ 0.008 & 0.857 $\pm$ 0.006 & 0.710 $\pm$ 0.009 \\
MC Dropout & 0.467 $\pm$ 0.014 & 0.709 $\pm$ 0.010 & 0.666 $\pm$ 0.016 & 0.499 $\pm$ 0.008 & 0.858 $\pm$ 0.006 & 0.717 $\pm$ 0.009 \\
SNGP & \textbf{0.448 $\pm$ 0.011} & \textbf{0.879 $\pm$ 0.003} & \textbf{0.845 $\pm$ 0.007} & \textbf{0.856 $\pm$ 0.004} & \textbf{0.961 $\pm$ 0.002} & \textbf{0.905 $\pm$ 0.005} \\
\hline
FID & 56.5 & 214.08 & 222.35 & 312.32 & 322.53 & 351.25 \\
Sampling Pop. & 80873 & 5000 & 100000 & 5000 & 2299 & 70044 \\
\hline
\end{tabular}
\end{table*}


\subsection{Training and Testing}
\label{sec: training}

All models were trained from scratch on \uma{Nvidia L40S} GPUs using a ResNet-18 backbone, with AdamW optimizer, cross-entropy loss and a multi-step learning rate scheduler initialized at $1\times10^{-3}$ and \uma{fine-tuned for accuracy.} \JJN{Deeper ResNets were considered but showed negligible gains. ResNet-18 was chosen for benchmarking due to its favorable trade-off between accuracy and efficiency.} Early stopping was used to prevent overfitting, and data augmentation was applied to improve generalization and robustness. For MC baseline, predictions from ten forward passes with active dropout \uma{(droprate 0.2)} were averaged. Both the Baseline and SNGP models were trained on the Acevedo (WBC) and Wong (amyloid plaque) datasets and evaluated for OOD performance on other datasets. 
\subsection{Evaluation}
\label{sec:evaluation}

OOD performance was evaluated using Area Under the Receiver Operating Characteristic curve (AUROC), which measures separation between ID and OOD samples based on uncertainty. Uncertainty was inferred from maximum softmax probability (MSP), with lower values indicating higher uncertainty. AUROC reflects the probability that an OOD sample is assigned higher uncertainty than an ID sample, where 1.0 denotes perfect discrimination and 0.5 random guessing. 

\section{Results}
\label{sec:results}

Tab.\ref{tab:results_acevedo} summarizes the OOD detection performance of all methods trained on the Acevedo dataset. SNGP achieved near-perfect OOD-AUROC across all external OOD datasets (0.97–1.00), while maintaining competitive in-distribution performance (Tab.\ref{tab:acevedo_indist}), indicating that improved uncertainty estimation does not come at the cost of classification accuracy.

Tab.\ref{tab:results_wong} presents the OOD detection results for models trained on the Wong dataset. Notably, performance on the Tang dataset was below 0.5 AUROC (worse than random), likely because the task in Tang is similar to Wong (FID 56.5) and shares the same class labels. Since both datasets represent amyloid plaques, distinguishing them as OOD is inherently challenging. To quantify dataset similarity, we report the Fréchet Inception Distance (FID), where lower values indicate greater semantic similarity.
\JJN{Several OOD comparisons are cross-task (e.g., WBC vs. colon vs. cardiac), representing coarse distribution shifts. OOD metrics in Tab.~\ref{tab:results_acevedo} \& \ref{tab:results_wong} use 1K ID/OOD samples per dataset, averaged over 10 seeds.}

\begin{table}[t]
\centering
\caption{In-distribution performance on Acevedo}
\label{tab:acevedo_indist}
\setlength{\tabcolsep}{3pt}
\renewcommand{\arraystretch}{1.0}
\begin{tabular}{lcccc}
\hline
\textbf{Method} & \textbf{Accuracy} & \textbf{ECE} & \textbf{F1} & \textbf{Latency (ms)} \\
\hline
Baseline & 0.984 & 0.004 & 0.982 & 0.191 \\
MC Dropout & 0.984 & 0.004 & 0.982 & 1.460 \\
SNGP & 0.981 & 0.003 & 0.979 & 0.213 \\
\hline
\end{tabular}
\end{table}

\begin{table}[t]
\centering
\caption{In-distribution performance on Wong}
\label{tab:wong_indist}
\setlength{\tabcolsep}{3pt}
\renewcommand{\arraystretch}{1.0}
\begin{tabular}{lcccc}
\hline
\textbf{Method} & \textbf{Accuracy} & \textbf{ECE} & \textbf{F1} & \textbf{Latency (ms)} \\
\hline
Baseline & 0.985 & 0.006 & 0.985 & 0.238 \\
MC Dropout & 0.985 & 0.005 & 0.985 & 1.502 \\
SNGP & 0.985 & 0.005 & 0.985 & 0.252 \\
\hline
\end{tabular}
\end{table}

\section{Conclusion}
\label{sec:conclusion}

SNGP provides an efficient and reliable framework for uncertainty-aware classification in biomedical imaging. Across multiple datasets, it maintains strong calibration and in-distribution accuracy while substantially improving OOD detection over deterministic and Monte Carlo methods. As a single-pass, lightweight modification to DNNs, SNGP enables safer deployment in new digital pathology settings. \JJN{While our study focuses on SNGP as a lightweight, uncertainty-aware baseline, there are other recent OOD methods that warrant future benchmarking in digital pathology.}

\section{Compliance with Ethical Standards and Acknowledgments}
\label{sec:ethics_acknowledgments}

This study used only publicly available, deidentified data from the MicroBench meta-dataset \cite{LozanoNirschl2024_microbench}. \JJN{This work was supported by the University of Wisconsin–Madison ADRC (NIH P30-AG53425), Department of Pathology, and the Madison VA GRECC}. The authors declare no conflicts of interest. AI tools were used only for grammar and spelling.



\bibliographystyle{IEEEbib}
\bibliography{strings,refs}

\begin{thebibliography}{10}

\bibitem{guo2017_dnn_calib}
Chuan Guo, Geoff Pleiss, Yu~Sun, and Kilian~Q. Weinberger,
\newblock ``On calibration of modern neural networks,''
\newblock in {\em Proceedings of the 34th International Conference on Machine Learning}. 2017, p. 1321–1330, JMLR.org.

\bibitem{gal2016_dropout_bayesian}
Yarin Gal and Zoubin Ghahramani,
\newblock ``Dropout as a bayesian approximation: Representing model uncertainty in deep learning,''
\newblock in {\em Proceedings of The 33rd International Conference on Machine Learning}, Maria Balcan and Kilian Weinberger, Eds., New York, New York, USA, 20--22 Jun 2016, vol.~48 of {\em Proceedings of Machine Learning Research}, pp. 1050--1059, PMLR.

\bibitem{lakshminarayanan2017_deep_ensembles}
Balaji Lakshminarayanan, Alexander Pritzel, and Charles Blundell,
\newblock ``Simple and scalable predictive uncertainty estimation using deep ensembles,''
\newblock in {\em Advances in Neural Information Processing Systems}, I.~Guyon, U.~Von Luxburg, S.~Bengio, H.~Wallach, R.~Fergus, S.~Vishwanathan, and R.~Garnett, Eds. 2017, vol.~30, Curran Associates, Inc.

\bibitem{Acevedo2020-ja}
Andrea Acevedo, Anna Merino, Santiago Alf{\'e}rez, {\'A}ngel Molina, Laura Bold{\'u}, and Jos{\'e} Rodellar,
\newblock ``A dataset of microscopic peripheral blood cell images for development of automatic recognition systems,''
\newblock {\em Data Brief}, vol. 30, no. 105474, pp. 105474, June 2020.

\bibitem{jung2022wbc}
Changhun Jung, Mohammed Abuhamad, David Mohaisen, Kyungja Han, and DaeHun Nyang,
\newblock ``Wbc image classification and generative models based on convolutional neural network,''
\newblock {\em BMC Medical Imaging}, vol. 22, no. 1, pp. 94, 2022.

\bibitem{Nirschl2018-pc}
Jeff Nirschl, Andrew Janowczyk, Eliot Peyster, Renee Frank, Ken Margulies, Michael Feldman, and Anant Madabhushi,
\newblock ``A deep-learning classifier identifies patients with clinical heart failure using whole-slide images of h\&e tissue,''
\newblock {\em PLoSOne}, vol. 13, no. 4, Apr. 2018.

\bibitem{Tang2019-fy}
Ziqi Tang, Kangway~V Chuang, Charles DeCarli, Lee-Way Jin, Laurel Beckett, Michael~J Keiser, and Brittany~N Dugger,
\newblock ``Interpretable classification of alzheimer's disease pathologies with a convolutional neural network pipeline,''
\newblock {\em Nat. Commun.}, vol. 10, no. 1, pp. 2173, May 2019.

\bibitem{Wong2022-nm}
Daniel~R Wong, Ziqi Tang, Nicholas~C Mew, Sakshi Das, Justin Athey, Kirsty~E McAleese, Julia~K Kofler, Margaret~E Flanagan, Ewa Borys, Charles~L White, 3rd, Atul~J Butte, Brittany~N Dugger, and Michael~J Keiser,
\newblock ``Deep learning from multiple experts improves identification of amyloid neuropathologies,''
\newblock {\em Acta Neuropathol. Commun.}, vol. 10, no. 1, pp. 66, Apr. 2022.

\bibitem{Kather2016-db}
Jakob~Nikolas Kather, Cleo-Aron Weis, Francesco Bianconi, Susanne~M Melchers, Lothar~R Schad, Timo Gaiser, Alexander Marx, and Frank~Gerrit Z{\"o}llner,
\newblock ``Multi-class texture analysis in colorectal cancer histology,''
\newblock {\em Sci. Rep.}, vol. 6, pp. 27988, June 2016.

\bibitem{LozanoNirschl2024_microbench}
Alejandro Lozano, Jeff Nirschl, James Burgess, Sanket~R Gupte, Yuhui Zhang, Alyssa Unell, and Serena Yeung-Levy,
\newblock ``$\mu$-bench: A vision-language benchmark for microscopy understanding,''
\newblock {\em NeurIPS}, vol. 38, 2024.

\bibitem{liu2020_sngp}
Jeremiah~Zhe Liu, Zi~Lin, Shreyas Padhy, Dustin Tran, Tania Bedrax-Weiss, and Balaji Lakshminarayanan,
\newblock ``Simple and principled uncertainty estimation with deterministic deep learning via distance awareness,'' 2020.

\end{thebibliography}

\end{document}